\documentclass{ecai}  % use option [doubleblind] for double blind submission and hiding the authors section

\usepackage{graphicx}
\usepackage{latexsym}
\usepackage{times}
\usepackage{soul}
\usepackage{url}
\usepackage[hidelinks]{hyperref}
\usepackage[utf8]{inputenc}
\usepackage{graphicx}
\usepackage{amsmath}
\usepackage{amsthm}
\usepackage{booktabs}
\usepackage{algorithm}
\usepackage{algorithmic}
\usepackage{amssymb}
\usepackage[normalem]{ulem}
\usepackage{float} 
\usepackage{subfigure}
\usepackage[title]{appendix}
\usepackage[pagewise]{lineno}

\begin{document}

\begin{frontmatter}
% Fine-grained Semantics Enhanced Contrastive Learning for Graphs
\title{Capturing Fine-grained Semantics in Contrastive Graph Representation Learning}

\author[A]{\fnms{Lin}~\snm{Shu}}
\author[A]{\fnms{Chuan}~\snm{Chen}\thanks{Corresponding Author. Email: chenchuan@mail.sysu.edu.cn.}}
\author[A]{\fnms{Zibin}~\snm{Zheng}} % use of \orcid{} is optional

\address[A]{Sun Yat-sen University}

\begin{abstract}
Graph contrastive learning defines a contrastive task to pull similar instances close and push dissimilar instances away. It learns discriminative node embeddings without supervised labels, which has aroused increasing attention in the past few years. Nevertheless, existing methods of graph contrastive learning ignore the differences between diverse semantics existed in graphs, which learn coarse-grained node embeddings and lead to sub-optimal performances on downstream tasks. To bridge this gap, we propose a novel \textbf{F}ine-grained \textbf{S}emantics enhanced \textbf{G}raph \textbf{C}ontrastive \textbf{L}earning (FSGCL) in this paper. Concretely, FSGCL first introduces a motif-based graph construction, which employs graph motifs to extract diverse semantics existed in graphs from the perspective of input data. Then, the semantic-level contrastive task is explored to further enhance the utilization of fine-grained semantics from the perspective of model training. Experiments on five real-world datasets demonstrate the superiority of our proposed FSGCL over state-of-the-art methods. To make the results reproducible, we will make our codes public on GitHub after this paper is accepted.
\end{abstract}

\end{frontmatter}

\section{Introduction}
Graph-structured data is pervasive in a wide variety of real-world scenarios, such as social graphs \cite{nettleton2013data,tan2019deep}, citation graphs \cite{asatani2018detecting,ebesu2017neural} and biological graphs \cite{muzio2021biological,jin2021application}. Recent years have witnessed the proliferation of network representation \cite{zhang2018network}, among which Graph Neural Networks (GNNs) \cite{wu2020comprehensive,zhou2020graph} attract a surge of interest and show their effectiveness in learning good-quality representations from graphs. However, most GNNs are trained under the supervision of manual labels \cite{kipf2016semi,velivckovic2017graph}, which are expensive and even unavailable in the real world, resulting in overfitting on the limited labels. As a consequence, self-supervised graph representation learning \cite{xie2022self,xu2021self} comes into being, which makes full use of unlabeled graphs to train models and thus removes the need for expensive manual labels.

\begin{figure}[htbp] %H为当前位置，!htb为忽略美学标准，htbp为浮动图形
\centering %图片居中
\includegraphics[width=0.5\textwidth]{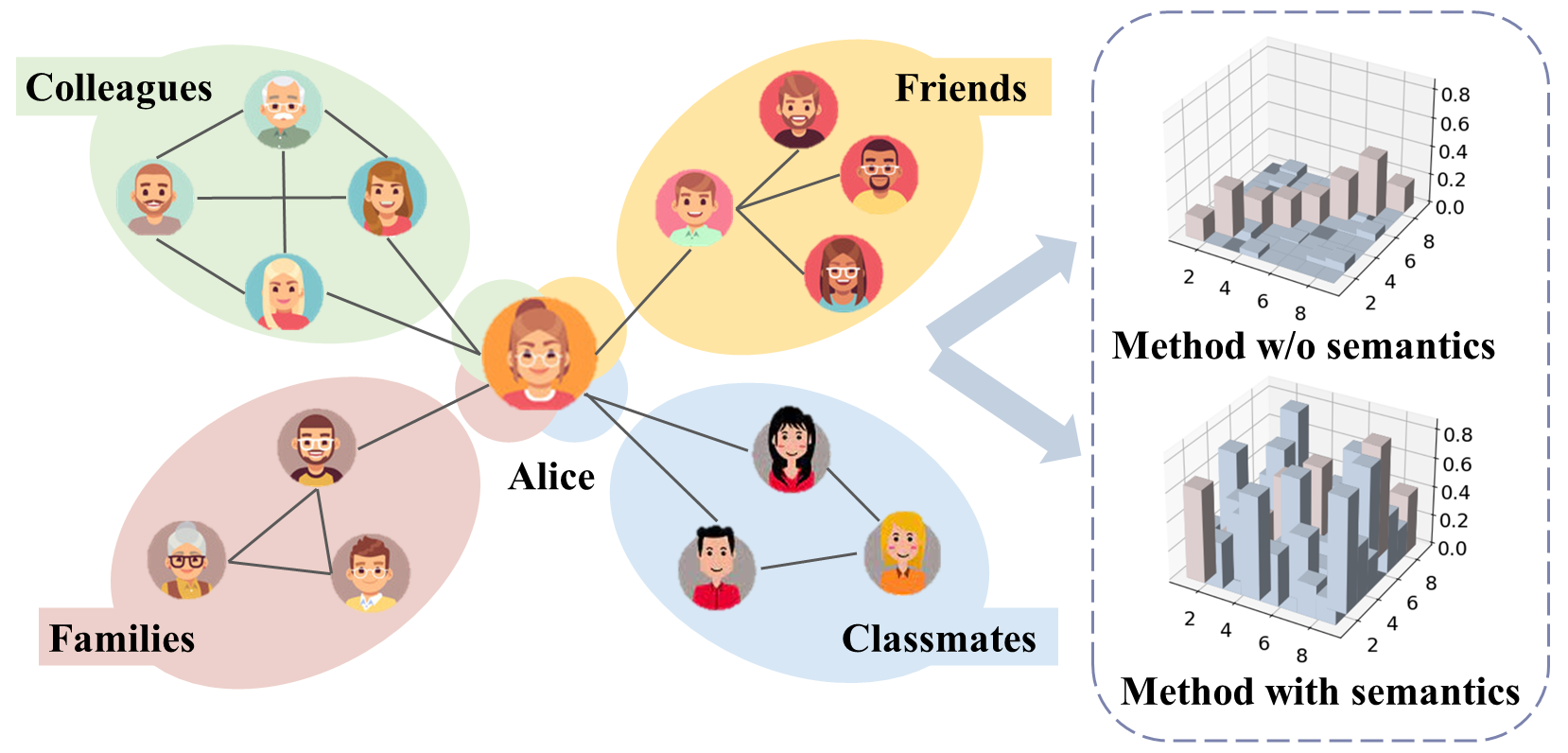} %插入图片，[]中设置图片大小，{}中是图片文件名
\caption{An illustrative example of diverse semantics existed in social graphs and the case study of utilizing fine-grained semantics in graph contrastive learning. The histograms depict the accuracies of node classification on the synthetic dataset.} %最终文档中希望显示的图片标题
\label{Fig.social_graph} %用于文内引用的标签
\end{figure}

As a dominative category of self-supervised graph representation learning, graph contrastive learning (GCL) \cite{you2020graph,zhu2021graph} has gained increasing attention in the past few years. Most GCL methods achieve an instance-level contrastive task, i.e., make each node/instance discriminative from each other. Nevertheless, there exist various latent semantics in graphs and different nodes might share high similarity under different inherent semantics. The instance-level contrastive schema constraints each node far away from each other, which ignores the differences between diverse semantics existed in graphs and is prone to pull nodes with high semantic similarity far away, resulting in coarse-grained representation for each node. As an illustration, Figure \ref{Fig.social_graph} depicts a social graph that contains four distinct communities. Specifically, Alice participates in four communities that imply diverse semantics of colleagues, friends, families and classmates. Existing GCL methods leverage the connections in graphs without distinguishing the semantic nuances between different connections, thus leading to the information loss of inherent semantics and sub-optimal performances on downstream tasks. To verify our hypothesis, we conduct a case study on a synthetic dataset composed of 8 communities, where nodes might belong to 2 communities simultaneously and each community can simulate the inherent semantics in graphs explicitly (More detailed settings are introduced in the Supplementary Material). Concretely, we apply the multilabel algorithm \cite{zhang2007ml} on the frozen node embeddings generated from \textit{method w/o semantics} (i.e., method that ignores fine-grained semantics \cite{lee2022augmentation}) and \textit{method with semantics} (i.e., our proposed method), where the ground-truth communities are regarded as labels. The histograms report the classification accuracies, where the value of the $i$-th row and $j$-th column denotes the proportion of nodes belonging to both community $i$ and $j$ that are accurately classified. As a result, the diagonal parts of histograms depict the accuracies of nodes with only one community, while the non-diagonal parts depict the accuracies of nodes with two communities. It can be observed that \textit{method w/o semantics} has lower classification accuracies on the non-diagonal parts compared with diagonal ones, implying that methods ignoring fine-grained semantics have poor learning ability on nodes with multiple semantics. In contrast, \textit{method with semantics} improves the accuracies of non-diagonal parts, indicating that considering the fine-grained semantics can boost the embedding quality and achieve higher performances on downstream tasks. Therefore, it is of curial significance to mine fine-grained semantics for graph contrastive learning.

In recent years, numerous studies \cite{biemann2016network,dareddy2019motif2vec} have verified the effectiveness of motifs for capturing latent semantics in graphs. Specifically, graph motifs are defined as essential subgraph patterns that occur frequently in graphs. Since different graph motifs depict different semantic contexts of each node, the semantic roles of each node can be easily distinguished according to the specified semantic contexts. As a consequence, graph motifs have been recognized as popular fundamental tools to mine latent semantics. For example, Motif-CNN \cite{sankar2017motif} fuses information from multiple graph motifs to mine semantic dependencies between nodes. SHNE \cite{zhang2021shne} captures structural semantics by exploiting graph motifs to boost the expressive power of the proposed model. Therefore, we propose to integrate graph motifs to mine fine-grained semantics for graph contrastive learning.

Overall, we propose \textbf{F}ine-grained \textbf{S}emantics enhanced \textbf{G}raph \textbf{C}ontrastive \textbf{L}earning (FSGCL) in this paper. Concretely, FSGCL first introduces motif-based graph construction, which generates multiple semantic graphs to integrate latent distinct semantics from the perspective of input data. Then, based on the generated semantic graphs, we combine the instance-level contrastive task with a carefully designed semantic-level contrastive task to further enhance the utilization of fine-grained latent semantics from the perspective of model training. In addition, since the random negative sampling strategy used in traditional GCL methods is prone to treat nodes with high similarity as negative pairs and pull them away, which results in the loss of semantic similarity, we adopt a slow-moving average approach \cite{grill2020bootstrap} to achieve the contrastive task without negative samples to further ensure the effective utilization of inherent semantics.

%Then, we follow the paradigm of graph contrastive learning, i.e., perform graph augmentations on both the original and constructed semantic graphs, and encode the input augmented graphs with semantic-wise graph encoder. Ultimately, the semantic-level contrastive task is designed to further enhance the utilization of fine-grained latent semantics from the perspective of model training. In addition, we adopt a slow-moving average approach \cite{grill2020bootstrap} to achieve the contrastive task without negative samples, which removes the need for random negative sampling and thus avoids pushing instances of similar semantics far away.

To sum up, the major contributions of this paper are concluded as follows:
\begin{itemize}
\item We propose a fine-grained semantics enhanced graph contrastive learning (FSGCL) model, which boosts the expressive ability of node representations by exploring latent semantics for contrastive learning on graphs. 
\item The motif-based graph construction is introduced to extract diverse semantics existed in graphs from the perspective of input data. Furthermore, the semantic-level and instance-level contrastive task without negative samples are explored jointly to further enhance the utilization of fine-grained semantics from the perspective of model training.
\item We conduct extensive experiments to demonstrate the effectiveness of FSGCL. We further report the results of visualization on the synthetic graph to verify the superior ability of capturing the diverse semantics of FSGCL.
\end{itemize}

\section{Related Work}
\subsection{Graph Contrastive Learning}
Graph contrastive learning has gained increasing popularity in recent years, which leverages unlabeled graphs to train models and thus removes the need for expensive manual labels. Inspired by the success of contrastive learning in computer vision \cite{dai2017contrastive,tian2020makes}, contrastive learning for graphs achieves the contrastive task by pulling similar instances close and pushing dissimilar instances far away. Representative researches include GraphCL \cite{you2020graph} which designs various types of graph augmentations to facilitate invariant representation learning for contrastive GNNs. Further, GCA \cite{zhu2021graph} proposes adaptive augmentation schemes to encourage the model to learn significant topologies and attributes from graphs. However, they exploit the connections in graphs without distinguishing the underlying semantic nuances, thus leading to sub-optimal node representations. Recently, PGCL \cite{lin2022prototypical} proposes a clustering-based approach that encourages semantically similar graphs closer and investigates the bias issue of negative sampling. DGCL \cite{li2021disentangled} disentangles the latent semantic factors of the graph to learn factorized graph representations for contrastive learning. Nevertheless, these methods are tailored for graph-level contrastive learning, which don't take the underlying node-level semantics into account and are not suitable for node-level graph contrastive learning.

\subsection{Motifs in Graph Learning}
Graph motif, a fundamental subgraph pattern that occurs frequently in graphs, plays a significant role in mining latent semantics. For instance, Motif-CNN \cite{sankar2017motif} fuses information from multiple graph motifs to mine semantic dependencies between nodes. SHNE \cite{zhang2021shne} captures structural semantics by exploiting graph motifs to boost the expressive power of the proposed model. However, these models are trained under the supervision of manual labels, which are expensive and even unavailable. Therefore, it is of crucial significance to utilize graph motifs to capture fine-grained semantics via graph contrastive learning, a dominative category of self-supervised graph representation learning. It should be noted that MICRO-Graph \cite{zhang2020motif} and MGSSL \cite{zhang2021motif} also incorporate graph motifs with self-supervised learning, but the goal of them is to generate informative motifs via self-supervised learning, where the latter model further leverages learned motifs to sample subgraphs for graph-level contrastive learning, whose aims are entirely different from that of our node-level graph contrastive learning model, which considers the underlying semantics to generate high-quality node embeddings.
\begin{figure}[htbp] %H为当前位置，!htb为忽略美学标准，htbp为浮动图形
\centering %图片居中
\includegraphics[width=0.35\textwidth]{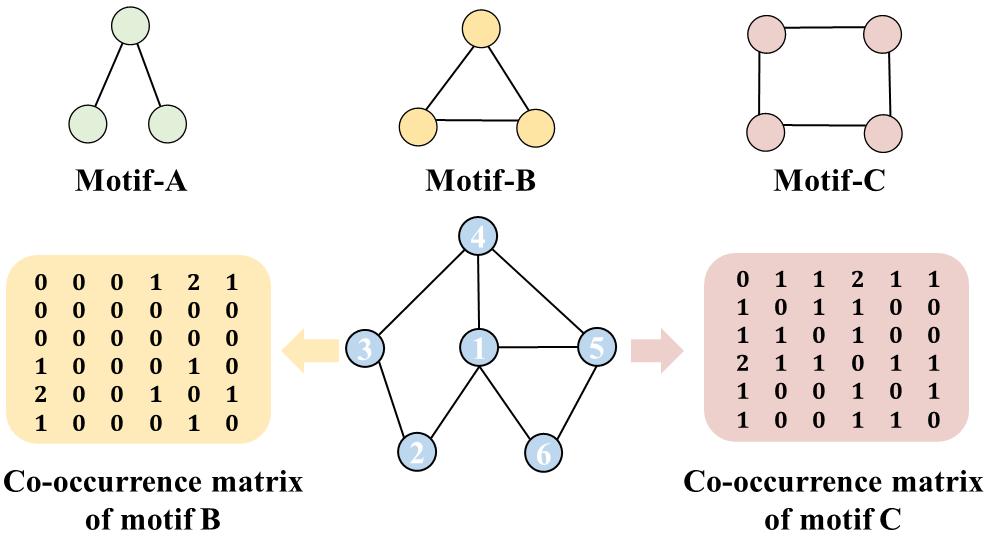} %插入图片，[]中设置图片大小，{}中是图片文件名
\caption{An illustrative example of graph motifs and corresponding co-occurrence matrices.} %最终文档中希望显示的图片标题
\label{Fig.motif} %用于文内引用的标签
\end{figure}

\section{Preliminaries}
In this section, we first describe the main definitions and notations that appeared in this paper. Specifically, calligraphic math font (e.g., $\mathcal{V}$) represents set, boldface upperclass letters (e.g., $\mathbf{A}$) represent matrices and boldface lowerclass letters (e.g., $\mathbf{w}$) represent vectors.

\textbf{Graph Contrastive Representation Learning.}
A graph can be represented as $\mathcal{G}=(\mathcal{V}, \mathcal{E})$, where $\mathcal{V}=\{v_1,v_2,...,v_n\}$ denotes the set of $n$ nodes and $\mathcal{E}$ denotes the set of links. Given a graph $\mathcal{G}$, graph contrastive representation learning aims at learning discriminative $d$-dimensional embeddings $\mathbf{Z}\in \mathbb{R}^{n\times d}$ for $n$ nodes in $\mathcal{G}$ via a contrastive task, which can be served for downstream tasks such as node classification. 

\textbf{Graph Motif.} A graph motif is denoted as $\mathcal{M}=(\mathcal{V}_\mathcal{M},\mathcal{E}_\mathcal{M})$, where $\mathcal{V}_\mathcal{M}$ and $\mathcal{E}_\mathcal{M}$ are the set of nodes and links of motif $\mathcal{M}$ respectively. Given a motif pattern $\mathcal{M}$, a \textit{motif instance} is defined as a subgraph of $\mathcal{G}$ that matches the pattern of motif $\mathcal{M}$, which is denoted as $(\mathcal{V}_S,\mathcal{E}_S)$, where $\mathcal{V}_S \subseteq \mathcal{V}$ and $\mathcal{E}_S \subseteq \mathcal{E}$, satisfying (1) $\forall u\in \mathcal{V}_S$, $\Phi (u)\in \mathcal{V}_\mathcal{M}$, where $\Phi:\mathcal{V}_S \rightarrow \mathcal{V}_\mathcal{M}$ is a bijection, (2) $\forall u,v \in \mathcal{V}_S$, if $(\Phi(u),\Phi(v))\in \mathcal{E}_\mathcal{M}$, then $(u,v)\in \mathcal{E}_S$. We denote the instance set of motif $\mathcal{M}$ as $S_{\mathcal{M}}=\{(\mathcal{V}_S,\mathcal{E}_S)\}$. 

\textbf{Motif-based Co-occurrence Matrix.} Given a specific motif $\mathcal{M}$ and corresponding instance set $S_{\mathcal{M}}=\{(\mathcal{V}_S,\mathcal{E}_S)\}$, the co-occurrence matrix $\mathbf{O}^\mathcal{M}$ of $\mathcal{M}$ is defined as:
\begin{equation}
\mathbf{O}^\mathcal{M}_{v,u}=\sum _{(\mathcal{V}_S,\mathcal{E}_S)} I((v,u)\in\mathcal{E}_S),
\end{equation}
where $I(s)$ is the indicative function, i.e., the value of $I(s)$ equals to 1 if the condition $s$ is true and 0 otherwise. $\mathbf{O}^\mathcal{M}_{v,u}$ depicts the number of co-occurrences of node $v$ and $u$ in all motif instances in $S_{\mathcal{M}}$. Figure \ref{Fig.motif} illustrates an example of graph motifs and corresponding co-occurrence matrices.

\begin{figure*}[htbp] %H为当前位置，!htb为忽略美学标准，htbp为浮动图形
\centering %图片居中
\includegraphics[width=0.93\textwidth]{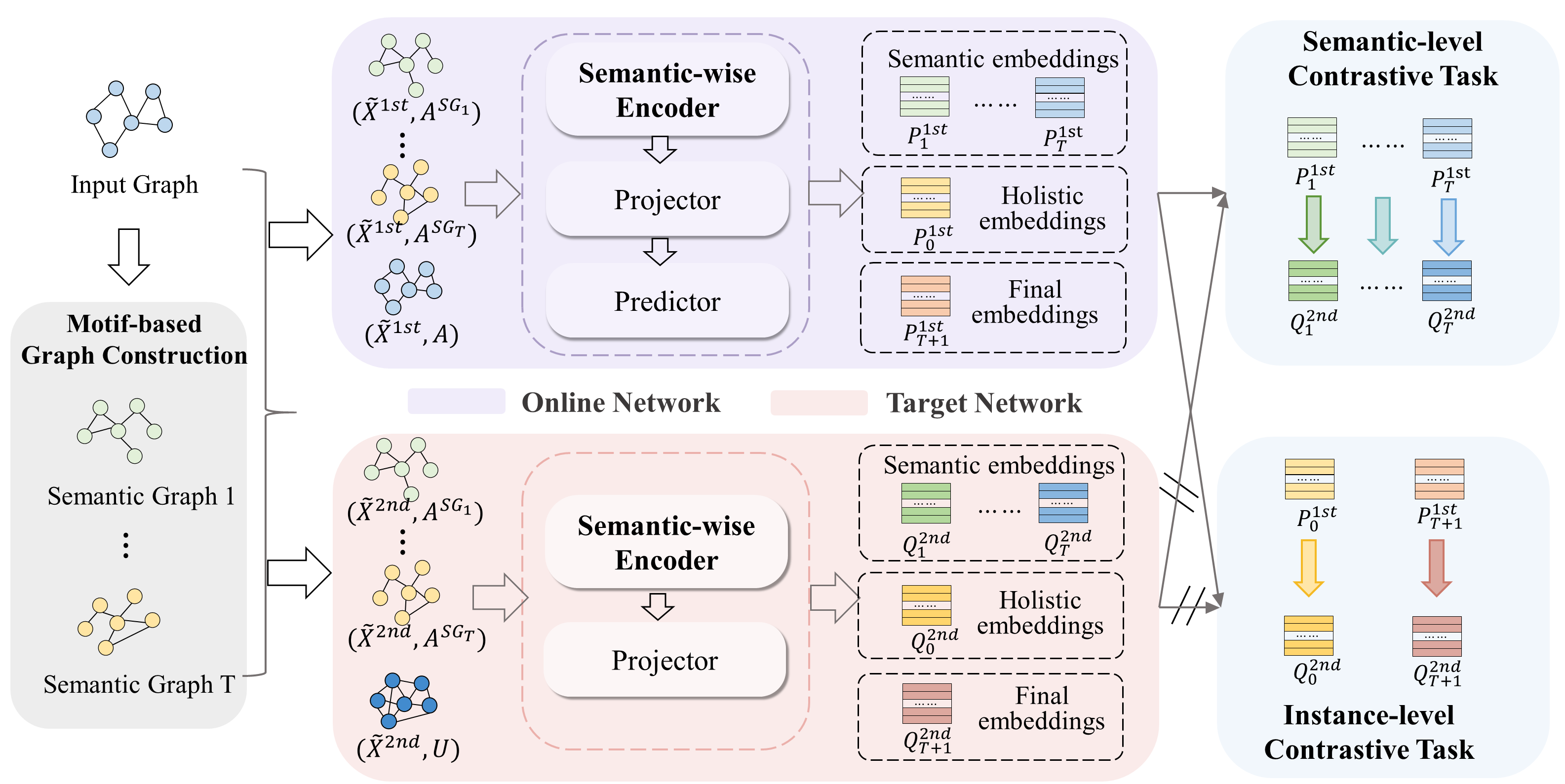} %插入图片，[]中设置图片大小，{}中是图片文件名
\caption{Framework of FSGCL. FSGCL utilizes the online and target network to learn, where the target network is stop-gradient and optimized by the slow-moving average of the online network.} %最终文档中希望显示的图片标题
\label{Fig.framework} %用于文内引用的标签
\end{figure*}
\section{Framework}
\subsection{Components of FSGCL}
Overall, FSGCL first generates two graph views $\mathcal{G}^{1st}$ and $\mathcal{G}^{2nd}$ based on the semantic graphs constructed by graph motifs (Section \ref{MotifSection}). Then, FSGCL utilizes two main neural networks to learn embeddings, i.e., the online network $f_{\theta}$ and target network $f_{\xi}\;(\theta\neq\xi)$. Concretely, the online network is composed of graph encoder, projector and predictor, while the target network has the same architecture as online network except for the absence of predictor. The asymmetric architecture prevents the training model from falling into trivial solutions. The semantic-wise graph encoder are introduced to extract latent semantics existed in graphs (Section \ref{SemanticSection}), followed by the semantic-level and instance-level contrastive objective formalized in Section \ref{ContrastiveSection}. The whole architecture of FSGCL is illustrated in Figure \ref{Fig.framework}.

\subsection{Motif-based Graph Construction}
\label{MotifSection}
To explicitly explore latent semantics in graphs, we leverage graph motifs to capture the semantic similarity. Specifically, given $T$ graph motifs, for each motif $\mathcal{M}_i,i\in\{1,...,T\}$, we first obtain the motif instance set $\mathcal{S_{M}}_i$ via an efficient motif matching algorithm \cite{lancichinetti2009benchmarks}, and calculate the corresponding motif-based co-occurrence matrix $\mathbf{O}^{\mathcal{M}_i}$, where node $u$ and $v$ share the same semantics within motif $\mathcal{M}_i$ if $\mathbf{O}^{\mathcal{M}_i}_{u,v}$ is non-zero. 
It is straightforward to employ the co-occurrence matrix as the semantic adjacency matrix to achieve the subsequent graph encoding procedure to generate node embeddings. However, on the one hand, graph motifs can only capture underlying semantics from the perspective of structural information, which ignore the role of node features played in semantic similarity. On the other hand, there exists a large number of similar node pairs in the co-occurrence matrix, thus it is unscalable and inflexible to regard all similar node pairs as neighbor pairs. As a consequence, we propose to facilitate both the feature similarity and motif co-occurrence matrix to select top-K neighbors for each node. Concretely, given the node feature $\mathbf{X}\in \mathbb{R}^{n\times d_f}$, where $d_f$ denotes the dimension of initial features, we first calculate the feature similarity matrix $\mathbf{C}$ through the widely used cosine similarity, which is formulated as:
\begin{equation}
\mathbf{C}_{ij}=\frac{\mathbf{X}_i \cdot \mathbf{X}_j}{\left \|\mathbf{X}_i \cdot \mathbf{X}_j \right \|}.
\end{equation}
Then, the non-zero position matrix of the co-occurrence matrix $\mathbf{O}^{\mathcal{M}_i}$ is extracted, which is denoted as $\mathbf{R}^{\mathcal{M}_i}$:
\begin{equation}
\mathbf{R}^{\mathcal{M}_i}_{u,v} = \left\{
	\begin{array}{lr}
	1, & \mathbf{O}^{\mathcal{M}_i}_{u,v}\neq 0, \\
	0, & \mathbf{O}^{\mathcal{M}_i}_{u,v}= 0.	
	\end{array}
\right.
\end{equation}
Ultimately, we select the top-K elements from $\mathbf{M}=\mathbf{C} \odot \mathbf{R}^{\mathcal{M}_i}$ for each node to construct the semantic graph $G^{SG_i}$, whose adjacency matrix $\mathbf{A}^{SG_i}$ is calculated as
\begin{equation}
\mathbf{A}^{SG_i}_{u,v} = \left\{
	\begin{array}{lr}
	\mathbf{M}_{u,v}, & \mathbf{M}_{u,v} \in TopK(\mathbf{M}_{u}), \\
	0, &  \mathbf{M}_{u,v} \notin TopK(\mathbf{M}_{u}),
	\end{array}
\right.
\end{equation}
where $\odot$ denotes the operation of Hadamard product. Therefore, we construct $T$ semantic graphs via $T$ graph motifs respectively, where the non-zero elements of each adjacency matrix $\mathbf{A}^{SG_i}$ denotes that two nodes share high semantic similarity within both motif $\mathcal{M}_i$ and node features.

Based on the semantic graphs, we generate two different graph views $G^{1st}$ and $G^{2nd}$ via graph augmentations, which help learn discriminative node embeddings for subsequent graph encoding and contrastive tasks. To further strengthen the information of the input graph, we collectively utilize the original graph and semantics graphs to build different graph views. Specifically, inspired by MVGRL \cite{hassani2020contrastive}, we first adopt the Personalized PageRank (PPR) \cite{page1999pagerank} diffusion on the input graph to compute the diffusion matrix $\mathbf{U}$:
\begin{equation}
\mathbf{U} = \alpha {(\mathbf{I}_n - (1-\alpha)\mathbf{D}^{-\frac{1}{2}}\mathbf{A}\mathbf{D}^{-\frac{1}{2}})}^{-1},
\end{equation}
where $\mathbf{A}$ is the adjacency matrix of the input graph, $\mathbf{D}\in \mathbb{R}^{n\times n}$ is the diagonal degree matrix of $\mathbf{A}$ and $\alpha$ denotes the teleport probability in a random walk \cite{klicpera2019diffusion}. Furthermore, we perturb the initial node feature matrix $\mathbf{X}$ to generate augmented features $\mathbf{\widetilde X}^{1st}$ and $\mathbf{\widetilde X}^{2nd}$ of different views, where each elements of $\mathbf{\widetilde X}^{1st}$ and $\mathbf{\widetilde X}^{2nd}$ is set to zero with a probability $r$. Overall, as depicted in Figure \ref{Fig.framework}, we build two graph views $G^{1st}$ and $G^{2nd}$ by 
\begin{equation}
\small
\left\{
	\begin{array}{lr}
	G^{1st}=\{(\mathbf{\widetilde X}^{1st},\mathbf{\widetilde A}^{1st}_i)| \mathbf{\widetilde A}^{1st}_{0}=\mathbf{A},\mathbf{\widetilde A}^{1st}_j=\mathbf{A}^{SG_j}\}, \\
	\\
	G^{2nd}=\{(\mathbf{\widetilde X}^{2nd},\mathbf{\widetilde A}^{2nd}_i)|\mathbf{\widetilde A}^{2nd}_{0}=\mathbf{U}, \mathbf{\widetilde A}^{2nd}_j=\mathbf{A}^{SG_j}\},
	\end{array}
\right.
\end{equation}
where $i\in\{0,...,T\}$ and $j\in\{1,...,T\}$. By applying graph diffusion and feature perturbation, we can not only simultaneously encode rich local and global information existed in the original graph \cite{hassani2020contrastive}, but also extract the intrinsic feature information. As a result, each graph view is composed of a holistic graph (i.e., the original whole graph) and $T$ semantic graphs that have been perturbed.

\subsection{Semantic-wise Graph Encoder}
\label{SemanticSection}
After generating two graph views based on the holistic graph and semantic graphs, we employ graph neural networks (GNNs) to learn node embeddings on each view. Since the $T+1$ graphs of each view depict distinct aspects of the input graph, encoding all graphs with the same GNN will result in the loss of unique information of each graph. Therefore, we propose to learn separate GNNs for each graph of each view, which are collectively referred to as \textit{semantic-wise graph encoder}. More specifically, for each graph $(\mathbf{\widetilde X}^m,\mathbf{\widetilde A}^m_i),i\in \{0,...,T\}$, where $m$ denotes the $m$-th augmented graph view, we utilize graph convolutional layers \cite{kipf2016semi} to learn node embeddings $\mathbf{Z}^m_i$, which update node embeddings by aggregating neighbors' embeddings. The propagation of each layer can be represented as:
\begin{equation}
\label{GCN}
\mathbf{Z}^{m}_i = \sigma (\mathbf{\widetilde{D}}_{i,m}^{-\frac{1}{2}}\mathbf{\overline A}^{m}_i \mathbf{\widetilde{D}}_{i,m}^{-\frac{1}{2}} \mathbf{Z}^{m}_i \mathbf{W}^{m}_i),
\end{equation}
where $\sigma$ is the activation function (i.e., PReLU), $\mathbf{\overline A}^{m}_i=\mathbf{\widetilde A}^{m}_i + \mathbf{I}_n$ is the adjacency matrix added with self-connections, $\mathbf{I}_n$ is the identity matrix, $\mathbf{\widetilde {D}}_{i,m}$ is the diagonal degree matrix of $\mathbf{\overline A}^{m}_i$, $\mathbf{W}_i^m \in \mathbb{R}^{d_{in} \times d_{out}}$ is the layer-specific transformation matrix for the $i$-th graph in the $m$-th view, $d_{in}$ and $d_{out}$ denote the input and output embedding dimensions of the propagation layer and $d_{out}$ is set to be $d$ in each propagation layer. Ultimately, the GNNs employed in semantic-wise graph encoder are built by stacking multiple propagation layers defined in Equation \ref{GCN}. For the convenience of presentation, we denote $\mathbf{Z}^{m}_{0}$ generated from the holistic graph as \textit{holistic embeddings} and $\mathbf{Z}^{m}_i, i\in\{1,...,T\}$ generated from semantic graphs as \textit{semantic embeddings} in the following sections.

Generally, for each augmented graph view, since different semantics emphasize distinct aspects of the whole graph, we apply the weighted sum function on $T$ semantic embeddings before combining them with the holistic embeddings to obtain the final node embeddings $\mathbf{Z}^m_{combine}$:
\begin{equation}
\label{weighted_sum}
\mathbf{Z}^m_{combine} = \beta \sum_{i=1}^{T} {w^m_i \mathbf{Z}^{m}_i} + \mathbf{Z}^{m}_{0},
\end{equation} 
where $w^m_i$ is the coefficient that balances different importance of $T$ semantics and $\beta$ is the weight coefficient that controls the balance between the holistic and semantic embeddings. As a consequence, the proposed semantic-wise graph encoder generates $T+2$ embeddings for each node $v$ on each graph view, which contains the holistic embeddings $\mathbf{Z}^{m}_{0}$, $T$ semantic embeddings $\mathbf{Z}^{m}_i, i\in\{1,...,T\}$, and the final node embeddings $\mathbf{Z}^{m}_{T+1}=\mathbf{Z}^m_{combine}$. 

%In the following sections, we collectively refer to the above $T+2$ embeddings as $\mathbf{Z}^{m}_i, i\in\{0,...,T+1\}$, where $\mathbf{Z}^{m}_{T+1}=\mathbf{Z}^m_{final}$.

\subsection{Contrastive Objective}
\label{ContrastiveSection}
The downstream task requires the output embeddings of graph encoders to learn, yet the optimization objectives of the downstream task and contrastive task are not consistent. As a consequence, before defining the contrastive objectives, we employ graph projectors to project the encoded embeddings $\mathbf{Z}_i^m$ to new subspaces, which have been confirmed to be an effective way to improve the representation quality \cite{chen2020simple}. Specifically, we leverage a non-linear perceptron layer to project node embeddings:
\begin{equation}
\mathbf{Q}^m_i = \sigma (\mathbf{U}_i^m \mathbf{Z}_i^m+b_i^m), i\in \{0,...,T+1\},
\end{equation}
where $\mathbf{U}_i^m \in \mathbb{R}^{d\times d}$ and $b_i^m\in \mathbb{R}^{d\times 1}$ are the transformation matrix and bias for the $i$-th embeddings of $m$-th graph view. Furthermore, for the online network, we stack $L$ perceptron layers as predictors to build an asymmetric architecture to prevent the training model from falling into trivial solutions:
\begin{equation}
\mathbf{P}_i^{1st} = \sigma (\mathbf{E}_i \mathbf{Q}_i^{1st} + a_i), i \in \{0,...,T+1\},
\end{equation}
where $\mathbf{E}_i \in \mathbb{R}^{d\times d}$ and $a_i\in \mathbb{R}^{d\times 1}$ are the transformation matrix and bias for the $i$-th embeddings in graph predictor.

In the following, we introduce the instance-level and semantic-level contrastive task in detail.
%we formalize the contrastive objectives by the semantic-level and holistic-level contrastive learning, which design contrastive tasks on the semantic graphs and holistic graph respectively.
\subsubsection{Instance-level Contrastive Learning}
The instance-level contrastive task regards the graph as a perceptual whole and achieves the contrastive task on the whole graph, i.e., it doesn't distinguish different semantics. In this section, we formalize the instance-level contrastive task on the holistic and final combined node embeddings since they don't distinguish diverse semantics. 

Traditional contrastive learning methods randomly select negative instances for each node to achieve the contrastive task. Nevertheless, random negative sampling introduces false-negative instance pairs, i.e., nodes that share similar semantics might be pulled far away, thus weakening the utilization of node similarity. Therefore, inspired by BYOL, we propose to design the instance-level contrastive task without negative samples. Specifically, we utilize the normalized predicted embeddings of online network $\mathbf{P}_i^{1st}$ to predict the normalized projected embeddings of target network $\mathbf{Q}_i^{2nd}$ and leverage the cosine similarity to calculate the loss:
%Similar to the semantic-level contrastive learning, we formalize the instance-level contrastive tasks via the predicting loss to further learn discriminative holistic and final combined node embeddings:
\begin{equation}
\small
L_{Holistic} = -\frac{1}{N}\sum_{v=1}^{N}{\frac{\mathbf{P}_{(0,v)}^{1st} \cdot \mathbf{Q}_{(0,v)}^{2nd}}{\left \| \mathbf{P}_{(0,v)}^{1st} \right \| \cdot \left \| \mathbf{Q}_{(0,v)}^{2nd} \right \|}},
\end{equation} 
\begin{equation}
\small
L_{Combine} = -\frac{1}{N}\sum_{v=1}^{N}{\frac{\mathbf{P}_{(T+1,v)}^{1st} \cdot \mathbf{Q}_{(T+1,v)}^{2nd}}{\left \| \mathbf{P}_{(T+1,v)}^{1st} \right \| \cdot \left \| \mathbf{Q}_{(T+1,v)}^{2nd} \right \|}},
\end{equation} 
where $\mathbf{P}_{(i,v)}^{1st}$ and $\mathbf{Q}_{(i,v)}^{2nd}$ denote node $v$'s embeddings of $\mathbf{P}_i^{1st}$ and $\mathbf{Q}_i^{2nd}$. As a result, the predicting loss only requires pushing pairs with high similarity closer.

\subsubsection{Semantic-level Contrastive Learning}
Since the instance-level contrastive learning ignores the fine-grained semantics existed in graphs, we propose the semantic-level contrastive task to strengthen the utilization of semantic information. Specifically, we leverage $T$ semantic embeddings to achieve the semantic-level contrastive learning. Since different semantic graphs convey distinct information, separate contrastive objectives should be designed on $T$ semantic graphs to distinguish semantic nuances. Therefore, for the $i$-th semantic graph, we formalize the semantic predicting loss without negative samples as
\begin{equation}
\small
L_{Semantic}^i = -\frac{1}{N}\sum_{v=1}^{N}{\frac{\mathbf{P}_{(i,v)}^{1st} \cdot \mathbf{Q}_{(i,v)}^{2nd}}{\left \| \mathbf{P}_{(i,v)}^{1st} \right \| \cdot \left \| \mathbf{Q}_{(i,v)}^{2nd} \right \|}}, i\in\{1,...,T\},
\label{semantic_loss}
\end{equation} 
where $\mathbf{P}_{(i,v)}^{1st}$ and $\mathbf{Q}_{(i,v)}^{2nd}$ denote node $v$'s embeddings of $\mathbf{P}_i^{1st}$ and $\mathbf{Q}_i^{2nd}$. The whole semantic-level contrastive loss is composed by the combination of $T$ semantic contrastive subtasks:
\begin{equation}
L_{Semantic} = \sum_{i=1}^T {L_{Semantic}^i}, i\in\{1,...,T\}.
\end{equation}

\renewcommand\arraystretch{1.4}
\begin{table*}
\small
\centering
\caption{Node classification accuracies ($\pm$ std) results (in $\%$) on five real-world datasets. OOM: out of memory on a 24GB GPU. We emphasize the highest values in \textbf{bold} and the second ones with \ul{underlines}. }
\begin{tabular}{c|c|c|c|c|c}
\hline
                & \textbf{Amazon-Photos}                  & \textbf{Amazon-Computers}              & \textbf{Coauthor-CS}                   & \textbf{Coauthor-Physics}              & \textbf{WikiCS}                        \\ \hline
\textbf{GCN}    & 93.18 $\pm$ 0.18          & 89.37 $\pm$ 0.29          & 92.34 $\pm$ 0.04          & 95.06 $\pm$ 0.06          & 79.19 $\pm$ 0.60          \\ \hline
\textbf{DGI}    & 91.61 $\pm$ 0.22          & 83.95 $\pm$ 0.47          & 92.15 $\pm$ 0.63          & 94.51 $\pm$ 0.52          & 75.35 $\pm$ 0.14          \\ \hline
\textbf{GMI}    & 90.68 $\pm$ 0.17          & 82.21 $\pm$ 0.31          & OOM                                    & OOM                                    & 74.85 $\pm$ 0.08          \\ \hline
\textbf{MVGRL}  & 91.74 $\pm$ 0.07          & 87.52 $\pm$ 0.11          & 92.11 $\pm$ 0.12          & 95.33 $\pm$ 0.03          & 77.52 $\pm$ 0.08          \\ \hline
\textbf{GRACE}  & 92.59 $\pm$ 0.16          & 88.55 $\pm$ 0.30          & 89.81 $\pm$ 0.19          & OOM                                    & 79.27 $\pm$ 0.67          \\ \hline
\textbf{GCA}    & 92.39 $\pm$ 0.32          & 87.57 $\pm$ 0.45          & 92.49 $\pm$ 0.14          & OOM                                    & 79.14 $\pm$ 0.25          \\ \hline
\textbf{BGRL}   & {\ul{93.23 $\pm$ 0.28}}    & {\ul{90.42 $\pm$ 0.15}}    & 92.68 $\pm$ 0.14            & 95.37 $\pm$ 0.10            & {\ul{80.07 $\pm$ 0.49}}    \\ \hline
\textbf{ProGCL} & 92.01 $\pm$ 0.24          & 87.44 $\pm$ 0.38          & OOM                                    & OOM                                    & 79.61 $\pm$ 0.63          \\ \hline
\textbf{AFGRL}  & 93.04 $\pm$ 0.26          & 89.41 $\pm$ 0.34            & {\ul{93.27 $\pm$ 0.17}}    & {\ul{95.71 $\pm$ 0.10}}     & 77.78 $\pm$ 0.42          \\ \hline
\textbf{FSGCL}    & \textbf{94.65 $\pm$ 0.16} & \textbf{90.54 $\pm$ 0.21} & \textbf{94.22 $\pm$ 0.07} & \textbf{96.10 $\pm$ 0.08} & \textbf{80.25 $\pm$ 0.02} \\ \hline

\end{tabular}
\label{results}
\end{table*}

\subsection{Model Training}
Combining the semantic-level and instance-level contrastive learning, we train the FSGCL model by the joint loss:
\begin{equation}
L_{Joint} = \gamma L_{Semantic} + L_{Holistic} + L_{Combine},
\end{equation}
where $\gamma$ is the weight coefficient that controls the significance of the semantic-level contrastive task. In practice, we symmetrize the loss by feeding $G^{2nd}$ and $G^{1st}$ to the online and target network respectively to compute $\widetilde L_{Joint}$. Furthermore, we only update the parameters $\theta$ of online network by minimizing the total loss $L=L_{Joint}+\widetilde L_{Joint}$, while the parameter $\xi$ of target network is stop-gradient and optimized by a slow-moving average of the online network:
\begin{equation}
\theta \leftarrow optimize(\theta,\lambda,\nabla_{\theta} L),
\end{equation}
\begin{equation}
\xi \leftarrow \tau\xi+(1-\tau)\theta,
\end{equation}
where $\lambda$ is the learning rate and $\tau$ is the decay rate. At the end of the training, we only keep the encoder of online network and treat the final node embeddings $\mathbf{Z}^{1st}_{final}$ as the input of downstream inference tasks.

Since the online and target network are initialized randomly, the randomness makes all node embeddings distinct from each other at the initial stage of training, thus achieving the effect of distinguishing different instances (i.e., the effect of negative instances). Moreover, $\tau$ is usually set to be a large number (e.g., 0.99), thus the slow-moving average algorithm can not only make augmented instances pairs (i.e., positive instance pairs) closer, but also implicitly preserve the distinction between different instances to a large extent. Compared to methods with random negative sampling, FSGCL only pulls positive instances closer, which not only reduces the computation cost, but also avoids pushing instances of similar semantics far away while optimization, thus strengthening the utilization of semantic similarity.

\section{Experiments}
\subsection{Datasets}
We conduct node classification on five public real-world datasets: Amazon-Photos, Amazon-Computers\footnote{\url{https://github.com/shchur/gnn-benchmark/tree/master/data/npz}}, Coauthor-CS, Coauthor-Physics\footnote{\url{https://github.com/shchur/gnn-benchmark/tree/master/data/npz}} and WikiCS\footnote{\url{https://github.com/pmernyei/wiki-cs-dataset/raw/master/dataset}}, where we use a random train/validation/test split (10/10/80$\%$) for the datasets of Amazon and Coauthor since these four datasets have no standard dataset splits, and follow the 20 canonical train/validation/test splits for WikiCS dataset. The details of datasets as listed in Table \ref{datasets}.

\begin{table}
\centering
\caption{Description of datasets and corresponding matching numbers of motif instances.}
\small
\tabcolsep=0.04cm
\begin{tabular}{c|ccccc}
\hline
\textbf{Datasets}  & \textbf{\begin{tabular}[c]{@{}c@{}}Amazon-\\ Photos\end{tabular}} & \textbf{\begin{tabular}[c]{@{}c@{}}Amazon-\\ Computers\end{tabular}} & \textbf{\begin{tabular}[c]{@{}c@{}}Coauthor-\\ CS\end{tabular}} & \textbf{\begin{tabular}[c]{@{}c@{}}Coauthor-\\ Physics\end{tabular}} & \textbf{WikiCS} \\ \hline
\textbf{Nodes}     & 7,650                                                              & 13,572                                                                & 18,333                                                           & 34,493                                                                & 11,701           \\
\textbf{Edges}     & 119,081                                                            & 245,861                                                               & 81,894                                                           & 247,962                                                               & 216,123          \\
\textbf{Attribute} & 754                                                               & 767                                                                  & 6,805                                                            & 8,415                                                                 & 300             \\
\textbf{Classes}   & 8                                                                 & 10                                                                   & 15                                                              & 5                                                                    & 10              \\ \hline
\textbf{Motif A}   & 12,136,593                                                          & 42,532,480                                                             & 1,409,965                                                         & 7,499,633                                                              & 36,875,788        \\
\textbf{Motif B}   & 717,400                                                            & 1,527,469                                                              & 85,799                                                           & 468,550                                                               & 3,224,375         \\
\textbf{Motif C}   & 14,516,676                                                          & 38,630,055                                                             & 202,620                                                          & 1,886,051                                                              & 79,649,534        \\ \hline
\end{tabular}
\label{datasets}
\end{table}

\subsection{Baselines and Experimental Settings}
We compare our FSGCL with semi-supervised GCN \cite{kipf2016semi} and 8 state-of-the-art unsupervised contrastive learning methods including DGI \cite{velickovic2019deep}, GMI \cite{peng2020graph}, MVGRL \cite{hassani2020contrastive}, GRACE \cite{zhu2020deep}, GCA \cite{zhu2021graph}, BGRL \cite{thakoor2021large}, ProGCL \cite{xia2022progcl}, AFGRL \cite{lee2022augmentation}. We use the authors' released codes and follow the papers' guidance to train all comparison methods. Since the three motif in Figure \ref{Fig.motif} are the most basic and simples motifs consisting of three nodes and four nodes,
we utilize three motifs shown in Figure \ref{Fig.motif} in this paper, which are denoted as \textit{Motif-A,B,C} in the following. More detailed experimental settings and parameter sensitivity analysis are presented in the Supplementary Material due to the limitation of pages.
% The proposed model FSGCL is implemented by PyTorch \cite{paszke2019pytorch} and DGL \cite{wang2019deep} with the AdamW optimizer.
% with the learning rate $0.001$ and weight decay of $10^{-5}$
%For node classification, the final evaluation is achieved by fitting an $l_2$-regularized LogisticRegression classifier from Scikit-Learn \cite{pedregosa2011scikit} using the liblinear solver on the frozened node embeddings, where the regularization strength is chosen by grid-search from $\{2^{-10},2^{-9},...,2^9,2^{10}\}$. The dimension of hidden and output representations are set to be 512.
%The number of $k, m, \beta, \gamma$ and layers of GCN and are set to be 7, 0.99, 0.5, 0.5, 2 for Amazon-Computers and WikiCS, while 5, 0.996, 1, 1, 1 for other datasets. The MLP layers $L$ of online predictor is set to be 2 for Amazon and Wiki datasets, and 1 for Coauthor datasets. The augmentation ratio $r$ is set to be 0.2 for Amazon-Computers, 0.1 for WikiCS and 0.3 for other datasets. As for the motif coefficient $w_i^m, i\in{motif-A,B,C}$, we set the value of $w^m$ to be $[0.7,0.1,0.2]$ for Amazon-Photos, $[0.3,0.1,0.6]$ for Amazon-Computers, $[0.4,0.2,0.4]$ for Coauthor-CS, $[0.5,0.45,0.05]$ for Coauthor-Physics and $[0.1,0.5,0.4]$ for WikiCS.

\subsection{Experimental Results}
We repeat all experiments 5 times and report the average accuracies ($\pm$ std) on five real-world datasets. The experimental results are reported in Table \ref{results}, where we emphasize the highest values in bold and the second ones with underlines. Overall, we summarize major observations as follows: 
\begin{itemize}
    \item Our proposed FSGCL consistently outperforms baselines of graph contrastive learning, which can be attributed to the ability of graph motifs that mine fine-grained semantics existed in graphs. Therefore, distinguishing nuances between diverse semantics in contrastive learning helps learn more informative node embeddings, thus boosting the performances of downstream tasks.
    \item It can be observed that FSGCL has large improvements on Amazon-Photos and Coauthor datasets, while the improvements are relatively small on Amazon-Computers and WikiCS. We conjecture that this phenomenon is because the matching numbers of motif instances are higher on Amazon-Computers and WikiCS than other datasets (shown in Table \ref{datasets}). To be more specific, we extract the non-zero node pairs in the motif co-occurrence matrices, and select top-K elements for each node based on the feature similarity of the aforementioned non-zero node pairs to construct semantic graphs. As a result, when there exist excess motif instances, a large amount of non-zero node pairs in the co-occurrence matrices are extracted, resulting in the adjacency matrices of semantic graphs being extremely close to the top-K matrices of feature similarity, thus having less informative semantics and achieving smaller improvements.
    \item Contrastive learning methods that don't require negative instances (i.e., our FSGCL, BGRL and AFGRL) perform better than those with negative sampling (i.e., DGI, GMI, MVGRL, GRACE, GCA and ProGCL). What's more, methods without negative sampling are more scalable and can easily be applied to larger datasets such as Coauthor-CS and Coauthor-Physics. It verifies that removing the procedure of negative sampling can not only reduce the computation cost, but also improve the embedding quality since it avoids pushing instances of similar semantics far away while optimizing.
\end{itemize}

\begin{figure}[tb]
\centering  %图片全局居中
\subfigure[ProGCL]{
\label{vi_ProGCL}
\includegraphics[width=0.21\textwidth]{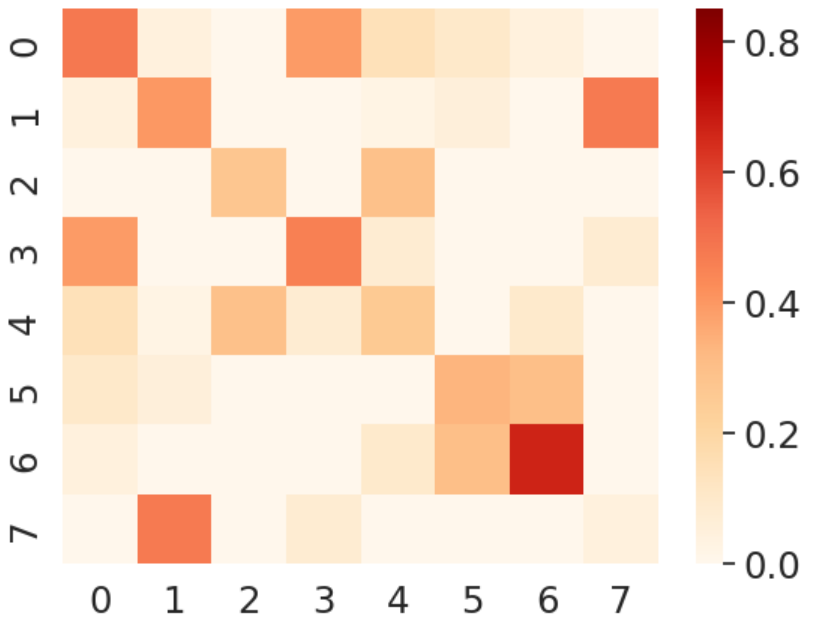}}
\subfigure[AFGRL]{
\label{vi_AFGRL}
\includegraphics[width=0.21\textwidth]{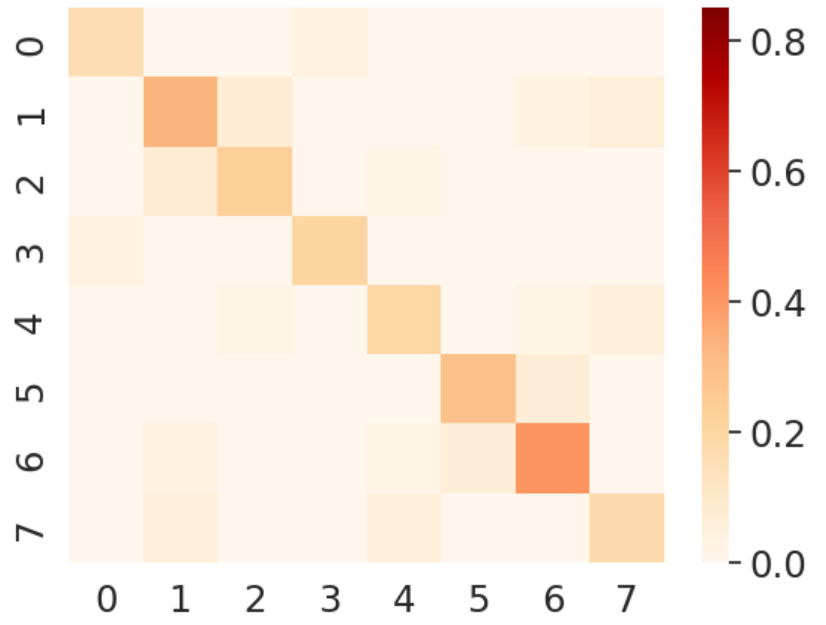}}
\\
\subfigure[BGRL]{
\label{vi_BGRL}
\includegraphics[width=0.21\textwidth]{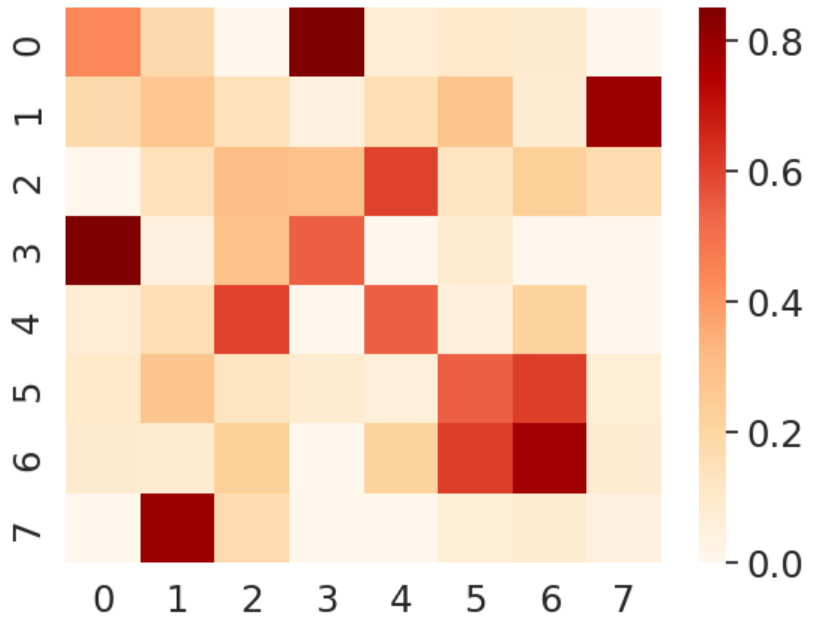}}
\subfigure[FSGCL]{
\label{vi_FSGCL}
\includegraphics[width=0.21\textwidth]{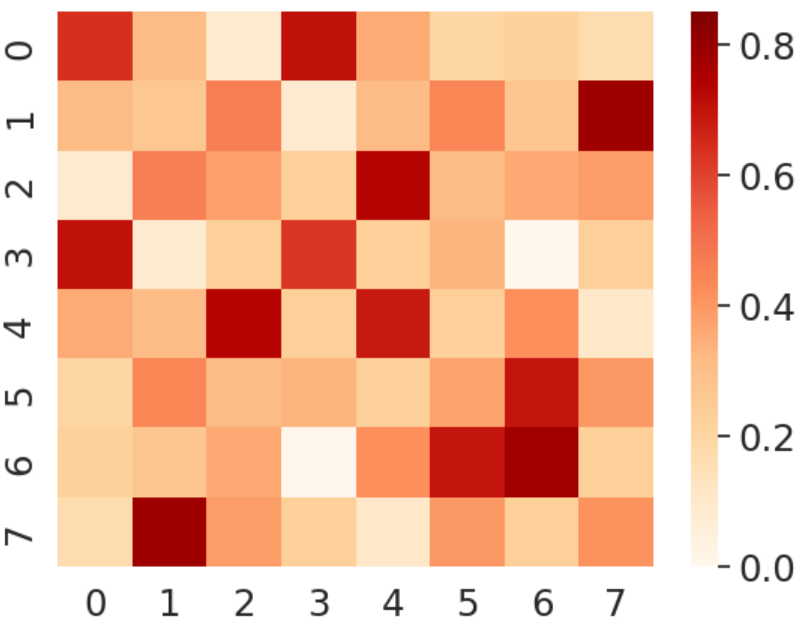}}
\caption{Visualization analysis on the synthetic datasets.}
\label{visualization}

\end{figure}

\subsection{Visualization Analysis on Synthetic Graphs}
To provide a more intuitive understanding of the functionality of mining fine-grained semantics of FSGCL, we employ the well-known LFR toolkit \cite{lancichinetti2009benchmarks} to generate a synthetic graph with overlapping communities, i.e., each node might exist in multiple communities, which can simulate the underlying diverse semantics more explicitly. Concretely, we generate a graph with 10,000 nodes and 8 communities, where 8,000 nodes simultaneously belong to two different communities (More detailed parameters are introduced in the Supplementary Material). Then, the multilabel K nearest neighbors algorithm \cite{zhang2007ml} is applied on the frozen node embeddings generated from ProGCL, AFGRL, BGRL and our FSGCL, where the ground-truth communities are used as labels.

Figure \ref{visualization} depicts the heatmaps of classification accuracies on 8 communities, where the value of the $i$-th row and $j$-th column denotes the proportion of nodes belonging to both community $i$ and $j$ that are accurately classified. Therefore, the diagonal elements of heatmaps imply the predicting accuracies of nodes which have only one community, while the non-diagonal elements imply the accuracies for nodes which have two distinct communities. It is apparent that ProGCL, AFGRL and BGRL have higher predicting accuracies on the diagonals compared with non-diagonal parts of heatmaps, which indicates that existing methods learn poor-quality embeddings for nodes that simultaneously participate in multiple communities since they ignore the distinct semantics existed in graphs. Instead, our FSGCL generates higher accuracies overall, especially on the non-diagonal parts of heatmaps, which demonstrates that FSGCL can indeed distinguish fine-grained semantics and boost the expressive ability.

\begin{table}
\centering
\small
\tabcolsep=0.01cm
\caption{Ablation study of key components of FSGCL.}
\begin{tabular}{l|ccccc}
\hline
\textbf{Datasets}  & \textbf{\begin{tabular}[c]{@{}c@{}}Amazon-\\ Photos\end{tabular}} & \textbf{\begin{tabular}[c]{@{}c@{}}Amazon-\\ Computers\end{tabular}} & \textbf{\begin{tabular}[c]{@{}c@{}}Coauthor-\\ CS\end{tabular}} & \textbf{\begin{tabular}[c]{@{}c@{}}Coauthor-\\ Physics\end{tabular}} & \textbf{WikiCS} \\ \hline
\textbf{FSGCL}     & \textbf{94.65}                                                             & \textbf{90.54}                                                                & \textbf{94.22}                                                           & \textbf{96.10}                                                                & \textbf{80.25}           \\ \hline
\textit{w/o\;$w_m^i$} & 94.07 & 90.45                                                                & 94.20 & 96.08 & 80.26           \\
\textit{w/o\;slow} & 92.20 & 89.80                                                                & 92.09 & 95.75 & 78.45           \\
\textit{w/o\; $\mathbf{A}^{SG_i}$} & 93.13                                                             & 90.05                                                                & 93.32                                                           & 95.75                                                                & 79.55           \\
\textit{w/o\;top-k\;$\mathbf{A}^{SG_i}$} & 93.97                                                                 & 89.96                                                                    & 94.10                                                               &                                                            96.03 & 78.97               \\
\textit{w/o\;$L_{Semantic}$} & 94.19                                                             & 90.49                                                                & 93.92                                                           & 96.02                                                                & 80.14           \\
\textit{w/o\;$L_{Holistic}$} & 94.15                                                             & 90.11                                                                & 93.87                                                           & 96.02                                                                & 79.60           \\ \hline
\end{tabular}
\label{ablation}
\end{table}

\subsection{Ablation Study}
%Variant 1: it doesn't utilize graph motifs to capture diverse semantics, i.e., this variant only adopts the non-negative training schema.
The ablation study is performed on six variants to investigate the effectiveness of key components:
\begin{itemize}
    \item \textit{w/o\;$w_i^m$}: it removes the weighted coefficients of different motifs in Eq.(\ref{weighted_sum}), i.e., it trains with an average weighted sum in Eq.(\ref{weighted_sum}).
    \item \textit{w/o\;slow}: it replaces the slow-moving average algorithm with the random negative sampling strategy to optimize the model.
    \item \textit{w/o\; $\mathbf{A}^{SG_i}$}: it sets $\mathbf{A}^{SG_i}=\mathbf{A}, i\in\{1,...,T\}$, i.e., it trains with original graphs instead of semantic graphs.
    \item \textit{w/o\;top-k\;$\mathbf{A}^{SG_i}$}: it sets $\mathbf{A}^{SG_i}=topK(\mathbf{C}), i\in\{1,...,T\}$, i.e., it substitutes the top-K matrices of feature similarity for the adjacency matrices of semantic graphs to train.
    \item \textit{w/o\;$L_{Semantic}$} and \textit{w/o\;$L_{Holistic}$}: they set $L_{Semantic}=0$ and $L_{Holistic}=0$, respectively.
\end{itemize}
The results of FSGCL and its variants are reported in Table \ref{ablation}. From this table, we can conclude that:
\begin{itemize}
    \item $w/o\;w_i^m$ performs slightly worse than FSGCL in most situations, demonstrating that considering different contributions made by distinct motifs helps improving the performances in downstream tasks. It should be noted that $w/o\;w_i^m$ is still superior than all comparison methods in Table \ref{results}, demonstrating that our FSGCL can achieve improvements with only few parameters.
    \item ${w/o\;slow}$ performs much worse than FSGCL, proving that the non-negative schema (i.e., the slow-moving average algorithm) can indeed improve the embedding quality by reducing the information loss of semantics. 
    \item We observe a performance drop in \textit{w/o\; $\mathbf{A}^{SG_i}$} and \textit{w/o\;top-k\;$\mathbf{A}^{SG_i}$}, verifying the effectiveness of graph motifs of capturing informative semantic nuances. 
    \item Compared with \textit{w/o\;$L_{Semantic}$} and \textit{w/o\;$L_{Holistic}$}, we can see that both semantic-level and instance-level contrastive loss can boost the performances of FSGCL, confirming the necessity of combining these two contrastive tasks.
\end{itemize}

\begin{table}
\centering
\small
\tabcolsep=0.02cm
\caption{Ablation study of different motif combinations of FSGCL.}
\begin{tabular}{l|ccccc}
\hline
\textbf{Datasets}  & \textbf{\begin{tabular}[c]{@{}c@{}}Amazon-\\ Photos\end{tabular}} & \textbf{\begin{tabular}[c]{@{}c@{}}Amazon-\\ Computers\end{tabular}} & \textbf{\begin{tabular}[c]{@{}c@{}}Coauthor-\\ CS\end{tabular}} & \textbf{\begin{tabular}[c]{@{}c@{}}Coauthor-\\ Physics\end{tabular}} & \textbf{WikiCS} \\ \hline
\textbf{FSGCL}     & \textbf{94.65}                                                             & \textbf{90.54}                                                                & \textbf{94.22}                                                           & \textbf{96.10}                                                                & \textbf{80.25}           \\ \hline
\textit{motif-A}   & 94.60                                                            & 90.51                                                               & 93.81                                                           & 96.02                                                                & 79.47           \\
\textit{motif-B}   & 93.74                                                            & 89.97                                                               & 93.49                                                           & 95.85                                                                & 79.89           \\
\textit{motif-C}   & 94.11                                                            & 90.21                                                               & 93.54                                                           & 95.85                                                                & 79.86           \\
\textit{motif-AB}  & 94.49                                                            & 90.50                                                               & 94.15                                                           & 96.07                                                                & 79.98           \\
\textit{motif-AC}  & 94.62                                                            & 90.52                                                               & 94.16                                                           & 96.05                                                                & 79.99           \\
\textit{motif-BC}  & 94.29                                                            & 94.30                                                               & 93.78                                                           & 95.83                                                                & 80.14           \\ \hline
\end{tabular}
\label{ablation_motif}
\end{table}

Besides, to investigate how graph motifs affect the performances of FSGCL, we conduct experiments on all combinations of motif-A,B,C used in this paper. From Table \ref{ablation_motif}, we can find that all combinations are worse than FSGCL which contains all three motifs, demonstrating that more graph motifs can help mine more semantic information and boost the expressive ability of node embeddings. It should be noted that our method can be seamlessly extended to utilize more complex graph motifs, we don't conduct extra experiments with more motifs in this paper.

\subsection{Runtime Analysis}
\begin{figure}[tb]
\centering  %图片全局居中
\subfigure[Amazon-Photos]{
\includegraphics[width=0.23\textwidth]{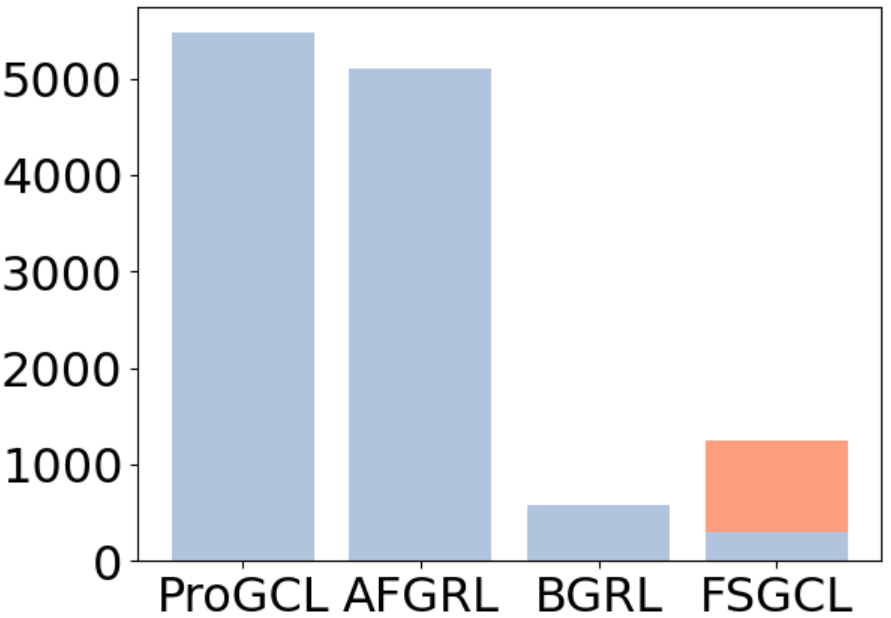}}
\subfigure[Amazon-Computers]{
\includegraphics[width=0.23\textwidth]{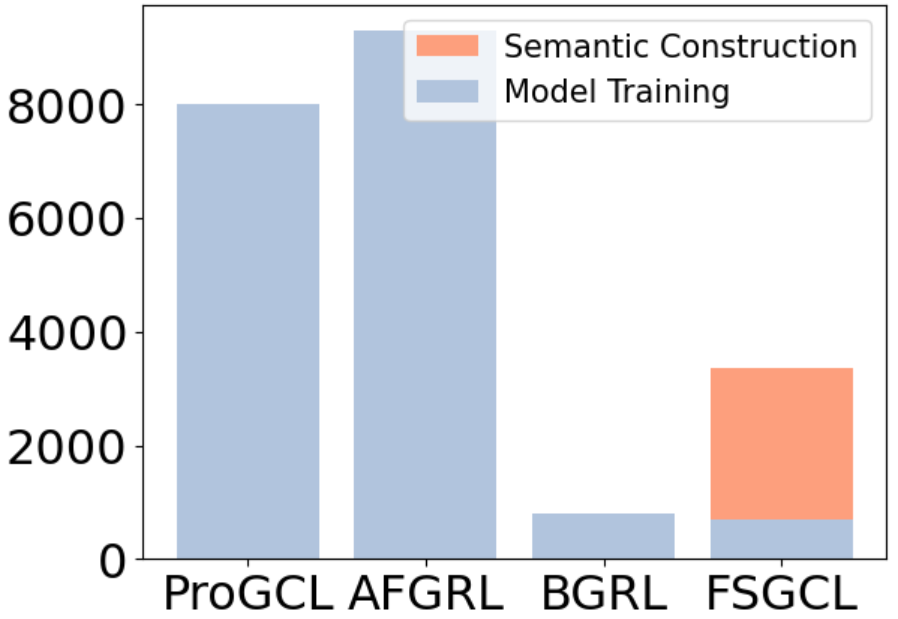}}
\caption{Runtime of FSGCL and three comparison methods on Amazon-Photos and Amazon-Computers.}
\label{runtime}

\end{figure}

In this section, we compare the runtime of our FSGCL and three methods (ProGCL, AFGRL and BGRL) that perform the best among all comparison methods. Specially, we not only report the model training time of FSGCL, but also report the time of semantic construction, i.e., the time of generating motif instances and constructing semantic graphs in the preprocessing process. The runtime results on Amazon-Photos and Amazon-Computers are depicted in Figure \ref{runtime}:
\begin{itemize}
    \item Compared with ProGCL requiring negative samples in model training, FSGCL reduces training time by removing the need of massive negative instances.
    \item FSGCL requices less training time than AFGRL. Though AFGRL also adopts the schema of non-negative training, this phenomenon is reasonable since AFGRL performs K-means clustering in each iteration, which is more time-consuming than our FSGCL that constructs semantic graphs during data preprocessing.
    \item Although FSGCL consumes more total time than BGRL, FSGCL owns less model training time than BGRL, which might because FSGCL converges faster by capturing fine-grained semantics and thus requiring fewer iterations and less training time.
\end{itemize}

%Therefore, when nodes simultaneously belong to multiple communities, methods that distinguish diverse semantics of each node can achieve more accurate classification results, thus having higher classification accuracies overall. In Figure \ref{visualization}, it can be observed that FSGCL generally has higher classification accuracy compared to other methods, verifying that our proposed method can indeed mine fine-grained latent semantics existed in graphs and boost the embedding quality.

\subsection{Conclusions}
In this paper, we propose a fine-grained semantics enhanced graph contrastive learning model - FSGCL. Concretely, FSGCL first employs graph motifs to construct multiple semantic graphs to mine the distinct semantics from the perspective of input data. Then, the semantic-level contrastive task without negative samples is introduced to further enhance the utilization of fine-grained latent semantics from the perspective of model training. The conducted experiments on five real-world datasets indicate that FSGCL is consistently superior to state-of-the-art methods, verifying the effectiveness of our proposed model.
\begin{appendices}
\begin{figure*}[ht]
\centering  %Í¼Æ¬È«¾Ö¾ÓÖÐ
\subfigure[Amazon Photos]{
\label{photo}
\includegraphics[width=0.3\textwidth]{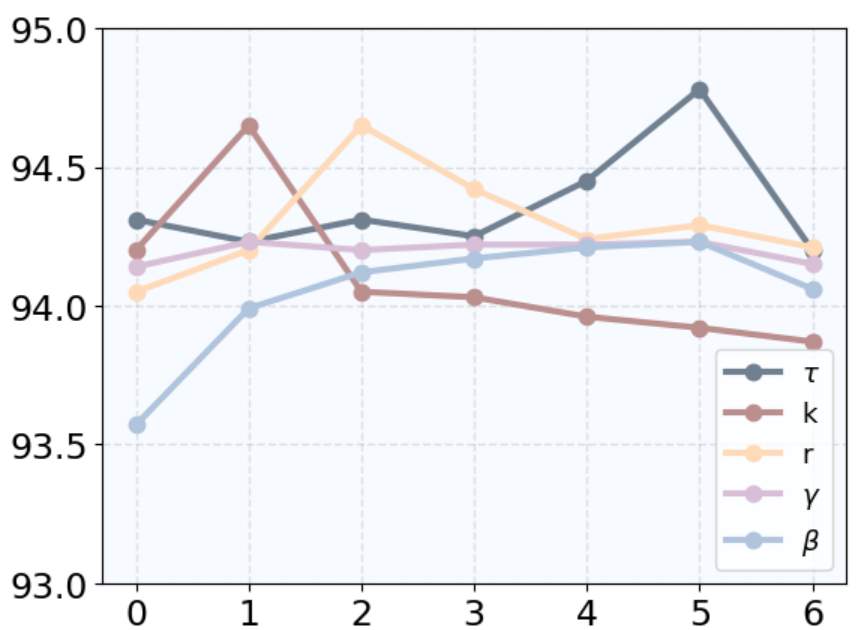}}
\subfigure[Amazon Computers]{
\label{computer}
\includegraphics[width=0.3\textwidth]{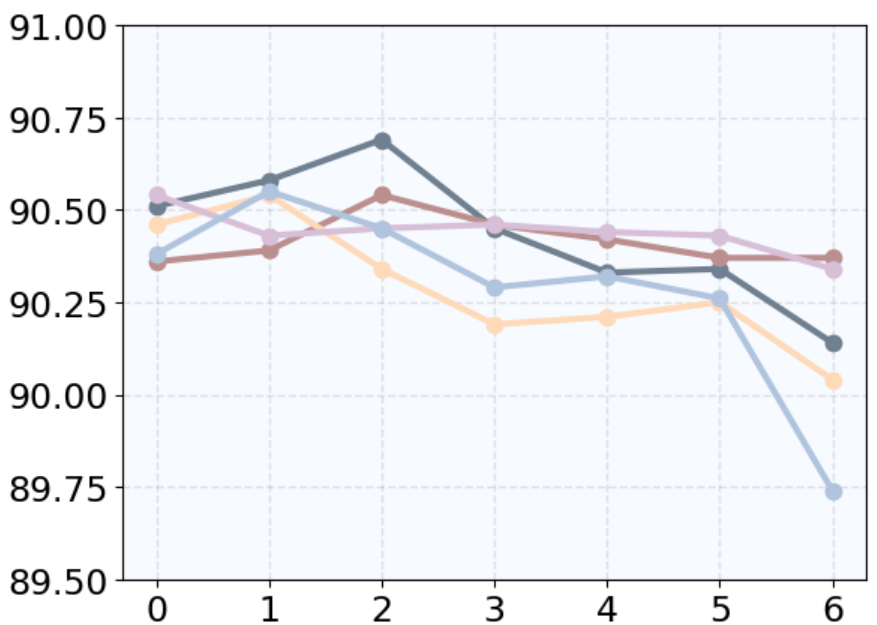}}
\\
\subfigure[Coauthor CS]{
\label{cs}
\includegraphics[width=0.3\textwidth]{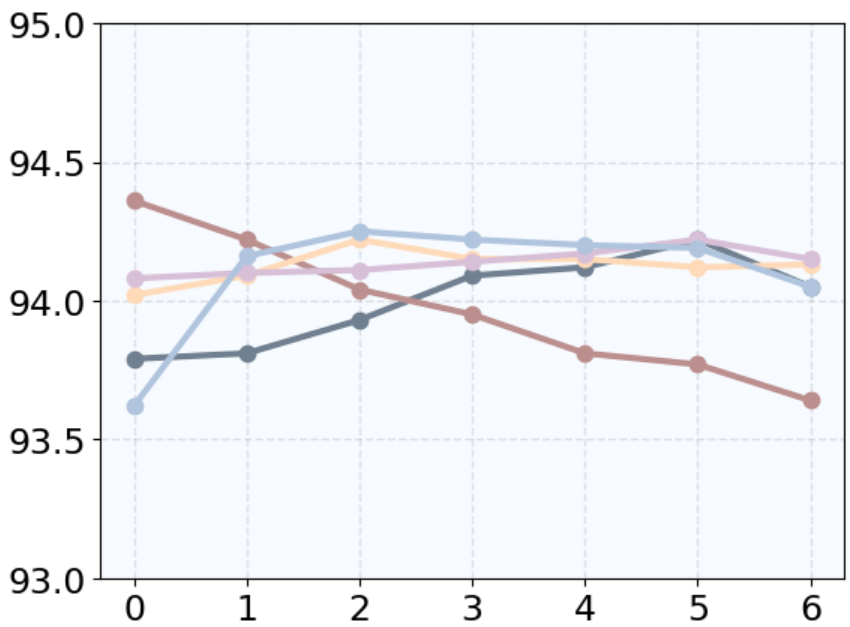}}
\subfigure[Coauthor Physics]{
\label{physics}
\includegraphics[width=0.3\textwidth]{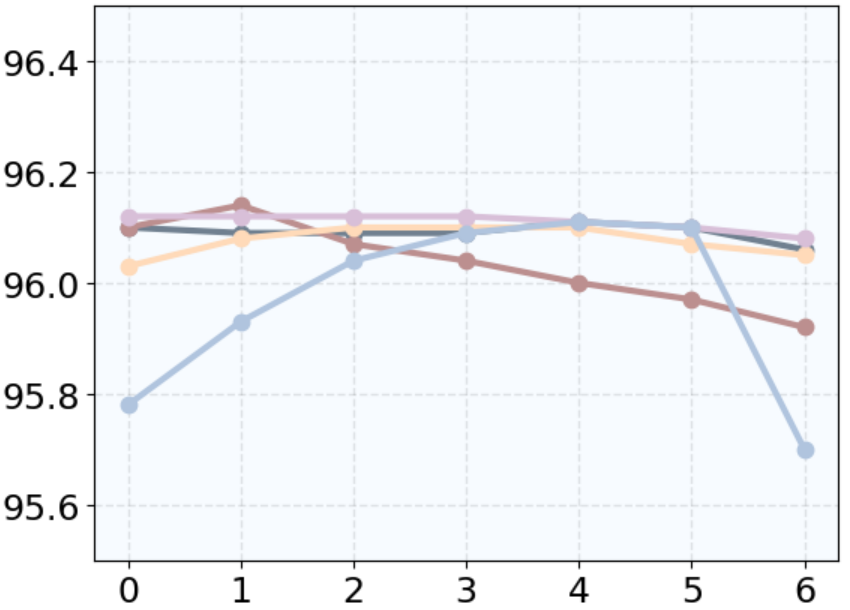}}
\subfigure[WikiCS]{
\label{wiki}
\includegraphics[width=0.3\textwidth]{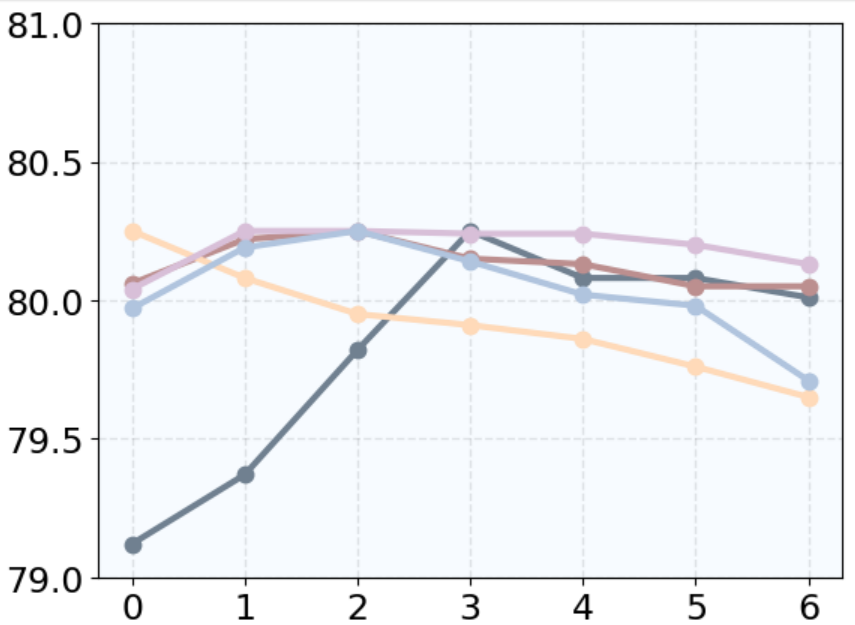}}
\caption{Parameter sensitivity analysis of $\tau, k, r, \gamma, \beta$ on five real-world datasets.}
\label{parameters}

\end{figure*}
\newpage

\section{Experimental Settings}
The proposed model FSGCL is implemented by PyTorch \cite{paszke2019pytorch} and DGL \cite{wang2019deep} with the AdamW optimizer with the base learning rate $\eta_{b}=0.001$ and weight decay of $10^{-5}$. Following the settings of BGRL \cite{thakoor2021large}, the learning rate is annealed via a cosine schedule   during the $n_{total}$ steps with an initial warmup period of $n_{w}$ steps. As a result, the learning rate at step $i$ is computed as:
\begin{equation}
\small
\eta_{i} = \left\{
	\begin{array}{lr}
	\frac{i\times \eta_{base}}{n_{w}}, & if\; i\leq n_{w} \\
	\\
	\eta_{b}\times(1+cos{\frac{(i-n_{w})\times \pi}{n_{total}-n_{w}}})\times 0.5, & if\;n_{w} \leq i \leq n_{total}
	\end{array}
\right.
\end{equation}

For node classification, the final evaluation is achieved by fitting an $l_2$-regularized LogisticRegression classifier from Scikit-Learn \cite{pedregosa2011scikit} using the liblinear solver on the frozen node embeddings, where the regularization strength is chosen by grid-search from $\{2^{-10},2^{-9},...,2^9,2^{10}\}$. The dimension of hidden and output representations are set to be 512. The number of $k, \tau, \beta, \gamma$ and layers of GCN are set to be 7, 0.99, 0.5, 0.5, 2 for Amazon-Computers and WikiCS, while 5, 0.996, 1, 1, 1 for other datasets. The MLP layers $L$ of online predictor is set to be 2 for Amazon and Wiki datasets, and 1 for Coauthor datasets. The augmentation ratio $r$ is set to be 0.2 for Amazon-Computers, 0.1 for WikiCS and 0.3 for other datasets. As for the motif coefficient $w_i^m, i\in{motif-A,B,C}$, we set the value of $w^m$ to be $[0.7,0.1,0.2]$ for Amazon-Photos, $[0.3,0.1,0.6]$ for Amazon-Computers, $[0.4,0.2,0.4]$ for Coauthor-CS, $[0.5,0.45,0.05]$ for Coauthor-Physics and $[0.1,0.5,0.4]$ for WikiCS. The value of $\alpha$ is set to be 0.2 on all datasets.

\section{Settings of Synthetic Dataset}
We employ the well-known LFR toolkit \cite{lancichinetti2009benchmarks} to generate a synthetic graph with 10,000 nodes and 8 overlapping communities. More specifically, we set the average degree $k$ and maximum degree to be 20 and 50 respectively, the mixing parameter $mu$ (each node shares a fraction of its edges with nodes in other communities) to be 0.2, the minimum number $minc$ and maximum number $maxc$ of community size to be 1500 and 3000, the number of overlapping nodes $on$ to be 8,000, and the number of memberships of the overlapping nodes $om$ to be 2.

\section{Parameter Sensitivity Analysis}

In this section, we investigate the sensitivity of five major hyperparameters: the decay rate $\tau$ of the slow-moving average strategy, $k$ that is used to select the top-K elements to build semantic graphs, the augmentation ratio $r$, the weight coefficient $\gamma$ that controls the significance of the semantic-level contrastive objective, and the weight coefficient $\beta$ that balances the holistic and semantic embeddings. The results of the above parameters are illustrated in Figure \ref{parameters}, where $\tau$ takes the value from the list  \{0.9,0.93,0.96,0.99,0.993,0.996,0.999\}, $k$ ranges in \{3,5,7,9,11,13,15\}, $r$ ranges in \{0.1,0.2,0.3,0.4,0.5,0.6,0.7\}, $\gamma$ and $\beta$ take the value from \{0.1,0.3,0.5,0.7,0.9,1,2\} and \{0.1,0.3,0.5,0.7,0.9,1,2\}, respectively.

As shown in Figure \ref{parameters}, FSGCL achieves higher performances when $\tau$ ranges from 0.99 to 0.999, demonstrating that a larger value of $\tau$ is beneficial for the slow-moving average strategy for learning discriminative embeddings. Besides, FSGCL always performs better when $k$ is small, which is reasonable since the semantic graphs own more unique semantic information when $k$ is smaller. For the augmentation ratio $r$, the performances drop sharply when $r$ becomes too large, indicating that over-perturbation leads to the loss of useful information in the original graphs. For the weight coefficient $\gamma$ and $\beta$, it can be observed that our model is relatively stable with regard to $\gamma$, while the performance decreases when $\beta$ is too small or too large.

\end{appendices}
\bibliography{ecai}
\end{document}